\documentclass{article}

    \PassOptionsToPackage{numbers, compress}{natbib}

     \usepackage[final]{neurips_2019}

\usepackage[utf8]{inputenc} 
\usepackage[T1]{fontenc}    
\usepackage{hyperref}       
\usepackage{url}            
\usepackage{booktabs}       
\usepackage{amsfonts}       
\usepackage{nicefrac}       
\usepackage{microtype}      
\usepackage{subcaption}
\usepackage{graphicx}
\usepackage{algorithm}
\usepackage{algpseudocode}
\usepackage{adjustbox}

\title{Federated and Differentially Private Learning for Electronic Health Records}

\author{%
  Stephen R. Pfohl\thanks{Work was done while an intern at Google Health} \\
  Stanford University \\
  \texttt{spfohl@stanford.edu} \\
   \And
   Andrew M. Dai \\
   Google Health \\
   \texttt{adai@google.com} \\
   \And
   Katherine Heller \\
   Google Health \\
   \texttt{kheller@google.com} \\
}

\begin{document}

\maketitle

\begin{abstract}
The use of collaborative and decentralized machine learning techniques such as federated learning have the potential to enable the development and deployment of clinical risk predictions models in low-resource settings without requiring sensitive data be shared or stored in a central repository.
This process necessitates communication of model weights or updates between collaborating entities, but it is unclear to what extent patient privacy is compromised as a result.
To gain insight into this question, we study the efficacy of centralized versus federated learning in both private and non-private settings. The clinical prediction tasks we consider are the prediction of prolonged length of stay and in-hospital mortality across thirty one hospitals in the eICU Collaborative Research Database. 
We find that while it is straightforward to apply differentially private stochastic gradient descent to achieve strong privacy bounds when training in a centralized setting, it is considerably more difficult to do so in the federated setting.
\end{abstract}

\section{Introduction}
The availability of high quality public clinical data sets \citep{johnson2016mimic,pollard2018eicu} has greatly accelerated research into the use of machine learning for the development of clinical decision support tools.
However, the majority of clinical data remain in private silos and are broadly unavailable for research due to concerns over patient privacy, inhibiting the collaborative development of high fidelity predictive models across institutions. 
Additionally, standard de-identification protocols provide limited safety guarantees against sophisticated re-identification attacks \citep{ElEmam2011a,gkoulalas2014publishing, kleppner2009committee}. 
Furthermore, patient privacy may be violated even in the case where no raw data is shared with downstream parties, as trained machine learning models are susceptible to membership inference attacks \citep{Shokri2017}, model inversion \cite{Fredrikson2015}, and training data extraction \cite{Carlini2018}.

In line with recent work \cite{Beaulieu-Jones2018,Vepakomma2018}, we investigate the extent to which several hospitals can collaboratively train clinical risk prediction models with formal privacy guarantees without sharing data. In particular, we employ \textit{federated averaging} \cite{McMahan2016} and \textit{differentially private stochastic gradient descent} \cite{McMahan2017,McMahan2018,Abadi2016} to train models for in-hospital mortality and prolonged length of stay prediction across thirty one hospitals in the eICU Collaborative Research Database (eICU-CRD) \cite{pollard2018eicu}. 

\subsection{Federated Learning}
Federated learning \cite{McMahan2016} is a general technique for decentralized optimization across a collection of entities without sharing data, typically employed for training machine learning models on mobile devices. 
In the variant known as \textit{federated averaging}, each entity trains a local model for a fixed number of epochs over the local training data and transfers the resulting weights to a central server. The server returns the average of the weights to each entity and the process repeats. 
This satisfies an intuitive notion of privacy, since no entity shares data with the central server or with any other entity. 
However, federated learning alone provides no formal accounting for the privacy cost incurred via the communication of local model weights with the central server.

\subsection{Differential Privacy}
Formally, a randomized algorithm $\mathcal{M}$: $\mathcal{D} \rightarrow \Theta$ with domain $\mathcal{D}$ and range $\Theta$ satisfies ($\epsilon$, $\delta$) differential privacy \cite{Dwork2014} if for any two adjacent data sets $d$, $d'$ $\in \mathcal{D}$ and for any subset of outputs $S \in \Theta$, 
\begin{equation}
    P(\mathcal{M}(d) \in S) \leq e^{\epsilon} P(\mathcal{M}(d') \in S) + \delta.
\end{equation}

In our case, the randomized algorithm $\mathcal{M}$ we consider is differentially private stochastic gradient descent (DP-SGD) \citep{Abadi2016,McMahan2018}. 
Here, adjacent data sets $d$, $d'$ are defined by adding, removing, and modifying the data for one record. 
This formulation can be informally interpreted as one where the inclusion of a record does not affect the probability distribution over learned model weights by more than a factor $\epsilon$, where $\delta$ bounds the probability of the restriction not holding.
Notably, this notion allows us to bound and quantify the capability for an adversary to determine whether a record belonged to the training data set, regardless of their access to auxiliary information \cite{Dwork2014}.

In practice, stochastic gradient descent can be made differentially private if the record-level gradients are clipped to a maximum $L_2$ norm $S$ and the Gaussian noise with standard deviation $\sigma$ added to the mean of the clipped gradients \cite{McMahan2018} over a batch of training data. 
The privacy loss over the procedure may then be accounted for with the moments accountant \cite{Abadi2016,McMahan2018} and Renyi differential privacy \citep{Mironov2017}. 
In this setting, the privacy cost of a training procedure is fully specified by the noise multiplier $\sigma / S$, the ratio of the batch size to the training set size, and the number of training steps \cite{McMahan2018}. 
\citet{McMahan2017} demonstrate that it is straightforward to formulate federated learning in a way that is conducive to differentially private training if DP-SGD is used as the local optimization algorithm.

\subsection{Related Work}
Our work is most similar to \citet{Beaulieu-Jones2018} in that they also investigate decentralized and differentially private machine learning in the context of mortality prediction in the context of the eICU-CRD, but use cyclical weight transfer \cite{Chang2018} rather than federated averaging for distributed optimization.
Another related technique is \textit{split learning} \cite{gupta2018distributed,Vepakomma2018a,vepakomma2019reducing} where the layers of a neural network are partitioned across several entities, enabling learning across entities that may contribute different data modalities without exposing the raw data or the local network architecture. 
As an alternative, recent work \cite{Beaulieu-Jones2017,Xie2018} has proposed the use of differentially private generative models to publicly release synthetic data with privacy guarantees.

\begin{table}
  \caption{
  Comparison of model performance for local, central, and federated training without differentially private training for each hospital and prediction task. Results shown are the estimates and 95\% confidence intervals for the AUC-ROC for local training and the relative difference in central and federated AUC-ROC compared to local training. Bold indicates a statistically significant improvement over local training on the basis of zero not being contained within the confidence interval for the difference in the AUC-ROC relative to the local model.
  }
  \label{tab:table_fed}
\centering
\begin{adjustbox}{width=0.95\textwidth}
\begin{tabular}{lllllll}
\toprule
{} & \multicolumn{3}{c}{Prolonged Length of Stay} & \multicolumn{3}{c}{Hospital Mortality} \\ \cmidrule(lr){2-4} \cmidrule(lr){5-7}
Hosp. ID (N) &           Local Abs. AUC-ROC &            Central Rel. AUC-ROC &          Federated Rel. AUC-ROC &           Local Abs. AUC-ROC &            Central Rel. AUC-ROC &          Federated Rel. AUC-ROC \\
\midrule
73 (4,381)  &  0.803 (0.761, 0.845) &   0.018 (-0.005, 0.042) &    \textbf{0.030 (0.012, 0.047)} &     0.791 (0.693, 0.890) &      \textbf{0.049 (0.007, 0.090)} &       0.025 (-0.007, 0.057) \\
264 (3,875) &  0.631 (0.571, 0.690) &    \textbf{0.058 (0.010, 0.106)} &    \textbf{0.067 (0.021, 0.114)} &     0.846 (0.785, 0.908) &     0.015 (-0.015, 0.046) &      -0.007 (-0.048, 0.034) \\
420 (3,167) &  0.707 (0.648, 0.766) &    \textbf{0.038 (0.003, 0.073)} &   0.016 (-0.014, 0.046) &     0.811 (0.729, 0.894) &     0.036 (-0.006, 0.079) &       0.019 (-0.017, 0.055) \\
338 (3,139) &  0.648 (0.584, 0.712) &    \textbf{0.098 (0.047, 0.149)} &    \textbf{0.105 (0.056, 0.154)} &     0.820 (0.722, 0.919) &     0.042 (-0.011, 0.094) &        \textbf{0.047 (0.008, 0.085)} \\
243 (3,026) &  0.664 (0.595, 0.732) &    \textbf{0.035 (0.003, 0.068)} &   0.022 (-0.008, 0.052) &     0.910 (0.864, 0.956) &    -0.006 (-0.035, 0.023) &      -0.025 (-0.066, 0.016) \\
458 (2,723) &  0.739 (0.679, 0.798) &    \textbf{0.042 (0.006, 0.078)} &    \textbf{0.039 (0.004, 0.075)} &     0.840 (0.754, 0.927) &     0.049 (-0.034, 0.131) &       0.030 (-0.034, 0.094) \\
167 (2,680) &  0.773 (0.715, 0.831) &   0.010 (-0.022, 0.042) &  -0.006 (-0.039, 0.028) &     0.845 (0.777, 0.914) &     0.031 (-0.016, 0.078) &       0.012 (-0.023, 0.048) \\
300 (2,678) &  0.718 (0.652, 0.783) &   0.040 (-0.002, 0.082) &   0.023 (-0.009, 0.055) &     0.699 (0.557, 0.841) &     0.051 (-0.014, 0.116) &       0.042 (-0.027, 0.112) \\
443 (2,666) &  0.700 (0.637, 0.763) &   0.037 (-0.002, 0.075) &    \textbf{0.052 (0.013, 0.090)} &     0.881 (0.816, 0.945) &     0.001 (-0.040, 0.043) &      -0.019 (-0.052, 0.013) \\
188 (2,591) &  0.773 (0.716, 0.830) &   0.014 (-0.021, 0.049) &   0.011 (-0.022, 0.044) &     0.850 (0.783, 0.917) &     0.013 (-0.041, 0.067) &      -0.003 (-0.062, 0.056) \\
208 (2,484) &  0.663 (0.595, 0.730) &    \textbf{0.090 (0.040, 0.140)} &    \textbf{0.044 (0.010, 0.077)} &     0.717 (0.596, 0.838) &      \textbf{0.114 (0.007, 0.220)} &       0.061 (-0.037, 0.159) \\
252 (2,449) &  0.802 (0.748, 0.855) &   0.026 (-0.011, 0.063) &   0.011 (-0.022, 0.044) &     0.829 (0.738, 0.920) &     0.049 (-0.024, 0.122) &       0.037 (-0.040, 0.113) \\
199 (2,215) &  0.760 (0.695, 0.825) &   0.016 (-0.024, 0.056) &   0.015 (-0.026, 0.055) &     0.838 (0.758, 0.918) &     0.033 (-0.023, 0.089) &       0.023 (-0.020, 0.066) \\
122 (2,103) &  0.681 (0.608, 0.755) &  -0.011 (-0.064, 0.042) &   0.001 (-0.044, 0.046) &     0.730 (0.605, 0.855) &     0.049 (-0.014, 0.112) &       0.045 (-0.010, 0.099) \\
176 (1,942) &  0.696 (0.618, 0.775) &    \textbf{0.066 (0.007, 0.124)} &    \textbf{0.056 (0.019, 0.092)} &     0.886 (0.819, 0.954) &     0.039 (-0.009, 0.086) &       0.039 (-0.003, 0.081) \\
281 (1,783) &  0.620 (0.528, 0.712) &    \textbf{0.169 (0.093, 0.244)} &    \textbf{0.084 (0.030, 0.137)} &     0.779 (0.605, 0.953) &      \textbf{0.101 (0.006, 0.195)} &        \textbf{0.129 (0.005, 0.252)} \\
411 (1,747) &  0.726 (0.647, 0.806) &  -0.011 (-0.057, 0.035) &  -0.027 (-0.067, 0.013) &     0.925 (0.875, 0.975) &    -0.039 (-0.107, 0.029) &      -0.010 (-0.048, 0.027) \\
413 (1,730) &  0.709 (0.623, 0.795) &   0.025 (-0.028, 0.078) &  -0.008 (-0.058, 0.043) &     0.809 (0.668, 0.951) &     0.062 (-0.045, 0.169) &      -0.146 (-0.303, 0.011) \\
449 (1,613) &  0.801 (0.730, 0.872) &    \textbf{0.051 (0.014, 0.088)} &    \textbf{0.052 (0.016, 0.088)} &     0.854 (0.777, 0.931) &     0.031 (-0.016, 0.077) &       0.014 (-0.019, 0.048) \\
394 (1509) &  0.747 (0.666, 0.828) &  -0.024 (-0.084, 0.037) &  -0.012 (-0.051, 0.027) &     0.896 (0.811, 0.981) &     0.027 (-0.044, 0.097) &      -0.010 (-0.054, 0.033) \\
283 (1,478) &  0.603 (0.508, 0.698) &    \textbf{0.084 (0.011, 0.156)} &    \textbf{0.074 (0.012, 0.136)} &     0.807 (0.684, 0.929) &     0.082 (-0.029, 0.192) &       0.038 (-0.056, 0.132) \\
307 (1,433) &  0.639 (0.543, 0.735) &   0.001 (-0.070, 0.072) &   0.025 (-0.034, 0.084) &     0.857 (0.750, 0.964) &     0.026 (-0.056, 0.108) &       0.029 (-0.041, 0.098) \\
331 (1,397) &  0.668 (0.499, 0.836) &   0.091 (-0.020, 0.203) &   0.065 (-0.007, 0.137) &     0.431 (0.204, 0.658) &      \textbf{0.409 (0.076, 0.742)} &       \textbf{0.308 (-0.028, 0.644)} \\
148 (1,386) &  0.791 (0.717, 0.866) &  -0.025 (-0.080, 0.029) &  -0.025 (-0.083, 0.032) &     0.815 (0.566, 1.000) &     0.062 (-0.102, 0.225) &       0.031 (-0.067, 0.130) \\
345 (1,372) &  0.688 (0.599, 0.778) &   0.054 (-0.020, 0.128) &   0.040 (-0.028, 0.109) &     0.769 (0.605, 0.934) &     0.086 (-0.066, 0.237) &       0.080 (-0.067, 0.227) \\
417 (1,369) &  0.716 (0.629, 0.804) &    \textbf{0.082 (0.018, 0.147)} &   0.054 (-0.007, 0.116) &     0.687 (0.518, 0.856) &      \textbf{0.189 (0.018, 0.360)} &       0.048 (-0.071, 0.167) \\
165 (1,336) &  0.635 (0.536, 0.735) &   0.024 (-0.043, 0.092) &   0.045 (-0.012, 0.101) &     0.595 (0.425, 0.765) &      \textbf{0.337 (0.186, 0.487)} &        \textbf{0.313 (0.175, 0.452)} \\
248 (1,334) &  0.730 (0.641, 0.819) &   0.009 (-0.048, 0.065) &   0.036 (-0.011, 0.083) &     0.777 (0.559, 0.995) &    -0.012 (-0.099, 0.076) &      -0.033 (-0.081, 0.016) \\
416 (1,330) &  0.739 (0.648, 0.830) &    \textbf{0.073 (0.012, 0.134)} &    \textbf{0.050 (0.004, 0.096)} &     0.675 (0.391, 0.959) &      \textbf{0.203 (0.003, 0.403)} &        \textbf{0.220 (0.048, 0.393)} \\
110 (1,305) &  0.673 (0.565, 0.781) &    \textbf{0.097 (0.001, 0.192)} &    \textbf{0.075 (0.004, 0.147)} &     0.947 (0.871, 1.000) &     0.008 (-0.017, 0.033) &      -0.028 (-0.079, 0.024) \\
183 (1,268) &  0.746 (0.656, 0.835) &   0.037 (-0.033, 0.108) &   0.024 (-0.045, 0.093) &     0.802 (0.627, 0.977) &    -0.018 (-0.126, 0.090) &      -0.002 (-0.059, 0.056) \\
\bottomrule
\end{tabular}

\end{adjustbox}
\end{table}
\section{Methods}
All experiments are based on data derived from the eICU Collaborative Research Database \cite{pollard2018eicu}, a freely and publicly available intensive care database containing data from 139,367 unique patients admitted between 2014 and 2015 to 208 unique hospitals. Each patient may have one or more recorded hospital admissions, each composed of one or more ICU stays. 

We make predictions at 24 hours into hospital admissions that last at least 24 hours. We assign binary outcome labels for in-hospital mortality and prolonged length of stay if the patient dies during the remainder of the hospital admission or if the admission last longer than 7 days, respectively.

To construct a training set for supervised learning, we first partition the set of admissions by hospital and then split the data within each hospital by patient such that 80\%, 10\%, and 10\% of the patients are used for training, validation, and testing, respectively. We allow for multiple hospital admissions per patient, but no patient exists in more than one partition within the same hospital. We retain all hospitals with greater than 1,000 hospital admissions in its corresponding training data set. This procedure produces a cohort of 65,509 labeled hospital admissions across 31 unique hospitals. The incidence of in-hospital mortality and prolonged length of stay in the aggregate population is 7.3\% and 34.4\%, respectively.

We construct a feature representation as a function of data recorded within each hospital stay up to 24 hours into the stay. We extract all lab orders, lab results, medication orders, diagnoses, and active treatments, as well as the patient age at admission, gender, ethnicity, unit type, and admission source. 
Lab results and age are binned into three and four bins, respectively.
We aggregate over time, assigning a one for each feature if it is observed anywhere in the admission prior to 24 hours and a zero otherwise.

For all supervised learning tasks, we consider only logistic regression and feedforward networks with one hidden layer. 
We perform model selection on the basis of the area under the receiver operating curve (AUC-ROC) evaluated on the corresponding validation set following a grid search over relevant hyperparameters. 
Model performance is reported as the 95\% confidence interval of the AUC-ROC on the corresponding test set derived via DeLong's Method \cite{DeLong1988}. 
We similarly derive confidence intervals for the difference in the AUC-ROC between models to facilitate model comparisons.\footnote{It should be noted that this procedure produces a \textit{confidence interval for the difference} in the AUC-ROC between models, taking into account the correlated nature of the predictions made by two models. 
}
The Adam \cite{kingma2014adam} optimizer is used in each case.

\subsection{Experimental Design}
We conduct a series of experiments designed to evaluate the relative benefits of centralized and federated learning, and the associated privacy costs, over learning using only local data at each hospital. We evaluate the following experimental conditions:
\begin{itemize}
    \item \textbf{Local training with no collaboration}. We identify a high performing model for each hospital using only data from that hospital following a grid search over learning rates, batch size, and hidden layer size if the model is a feedforward network. 
    \item \textbf{Centralized training}. 
    We simulate the setting where all of the records are available in a central repository, selecting the best global model on the basis of the performance on the aggregated records and evaluate the model on the local data from each hospital. 
    \item \textbf{Centralized training with differential privacy}. 
    We modify the centralized training procedure to use DP-SGD for optimization \cite{McMahan2018}. Here we additionally search over the discrete grid of [0.1, 1, 10] for both the noise multiplier $\sigma / S$ and the gradient clipping threshold $S$. 
    We assess privacy in terms of the $\epsilon$ that results from training with a fixed $\delta = 10^{-5}$.
    \item \textbf{Federated learning}. 
    We employ the federated averaging algorithm described in \citet{McMahan2016}. 
    For each round of federated learning, we conduct one epoch of training using the local data at each hospital and then synchronize the weights across all hospitals with an average. 
    We maintain a record of the local performance at each hospital over the federated learning procedure and perform local model selection on the basis of the best validation AUC-ROC observed over the procedure. 
    Model selection for the best federated hyperparameters is determined on the basis of the best mean local validation AUC-ROC across hospitals.
    \item \textbf{Federated learning with differential privacy}. We repeat the federated averaging experiment as previously described, but use DP-SGD as the local optimizer at each hospital, similar to the algorithm described in \citet{McMahan2017}. We experiment with fixed global DP-SGD hyperparameters and with local hyperparameters selected independently at each hospital. For the local hyperparameter search at each hospital, we use $\delta= 10^{-5}$, $\sigma / S = 1$, and $S$ selected log uniformly from $[0.1, 1, 10]$, performing model selection on the basis of the DP-SGD hyperparameters that maximize local AUC-ROC in ten epochs of training without any collaboration. We then perform federated learning for ten rounds with the selected local DP-SGD hyper-parameters.
\end{itemize}

\section{Results and Discussion}
Prior to experimentation with differentially private training, we aimed to establish the efficacy of federated learning over centralized and local learning. We find that while there is often a benefit to federated learning over local learning, often attaining an AUC-ROC comparable with that of centralized learning, the improvements are often not large enough to be rendered statistically significant on the basis of the 95\% confidence interval for the difference in AUC-ROC between either the central or federated model with the corresponding local model (Table \ref{tab:table_fed}). 
In particular, centralized and federated learning for prediction of prolonged length of stay improve on local learning for thirteen and twelve hospitals, respectively, whereas centralized and federated learning only benefit mortality prediction in seven and five cases, respectively. 

\begin{figure}[t]
    \centering
    \begin{subfigure}[t]{0.45\textwidth}
        \centering
        \includegraphics[width=0.95\textwidth]{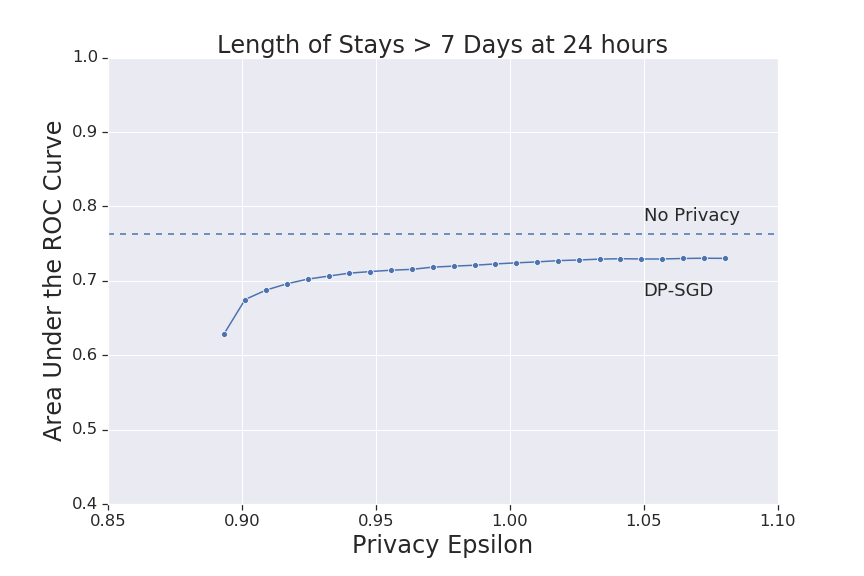}
    \end{subfigure}%
    ~ 
    \begin{subfigure}[t]{0.45\textwidth}
        \centering
        \includegraphics[width=0.95\textwidth]{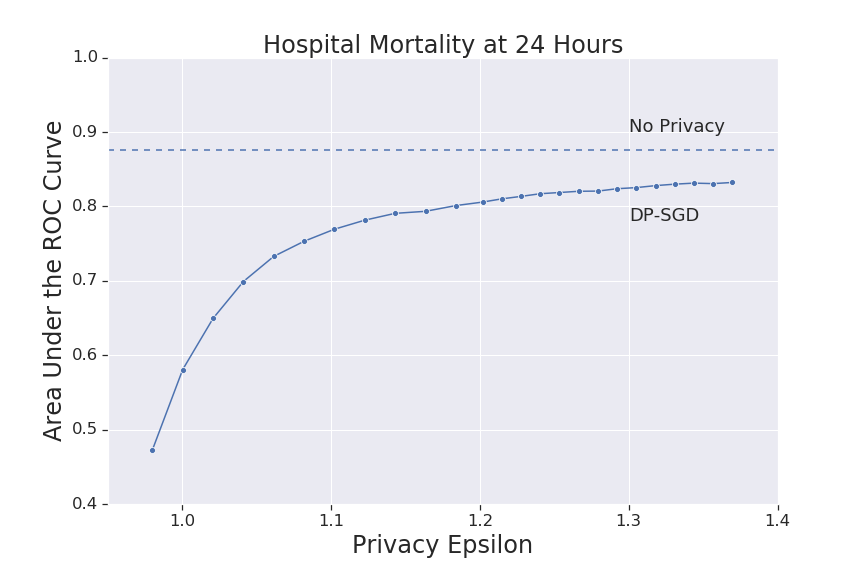}
    \end{subfigure}
    \caption{Trade-off between the differential privacy $\epsilon$ and validation AUC-ROC for $\sigma / S = 1.0$, $S = 10.0$ over a training procedure of 25 epochs.}
    \label{fig:central_dp}
\end{figure}

When the records from all hospitals are aggregated for differentially private centralized training, it is feasible to attain relatively strong privacy guarantees ($\epsilon \approx 1$) if $\sigma / S = 1.0$ and $S = 10.0$ (Figure \ref{fig:central_dp}) with a relatively minor reduction in terms of the validation AUC-ROC at the end of training (prolonged length of stay 0.763 vs. 0.73; mortality 0.876 vs. 0.832). When attempting to perform federated learning in a differentially private manner, we find that even with DP-SGD hyperparameters selected on the basis of local training, the models derived from differentially private federated learning often perform poorly in terms of both AUC-ROC and $\epsilon$, and that this effect is exacerbated for mortality prediction (Table \ref{tab:table_fed_dp}). It is likely that a practical tuning strategy for differentially private federated averaging could be identified with further experimentation, but it is unclear if such a strategy would generalize to similar data sets and prediction tasks. This is problematic, for both this and related work, as neglecting to account for the privacy cost of model selection produces optimistic underestimates of the privacy costs \cite{Liu2018a,chaudhuri2013stability}. In future work, it is of interest to conduct controlled experiments to directly compare our approach to cyclical weight transfer \cite{Beaulieu-Jones2018} and split learning \cite{gupta2018distributed,Vepakomma2018a,vepakomma2019reducing} to gain insight into the relative efficacy of differentially private federated averaging over alternatives.

\subsubsection*{Acknowledgments}
We thank Michaela Hardt and Abhradeep Thakurta for valuable mentorship and feedback. We further thank Steve Chien and all contributors to the Tensorflow Privacy project for enabling this work.

\bibliographystyle{unsrtnat}
\bibliography{references} 

\newpage
\appendix
\setcounter{table}{0}
\renewcommand{\thetable}{S\arabic{table}}

\begin{table}
  \caption{
  Model performance and privacy $\epsilon$ attained at each hospital following federated differentially private learning. Results shown are the best AUC-ROC on the validation set achieved over with early-stopping over ten rounds of federated learning. Prior to federated learning, the DP-SGD hyperparameters are selected separately at each hospital to maximize the local AUC-ROC after ten epochs of local differentially private training. 
  }
  \label{tab:table_fed_dp}
\centering
\begin{tabular}{rrrrrr}
\toprule
{} & {} & \multicolumn{2}{c}{Prolonged Length of Stay} & \multicolumn{2}{c}{Hospital Mortality} \\ \cmidrule(lr){3-4} \cmidrule(lr){5-6}
Hospital ID & N & \multicolumn{1}{c}{AUC-ROC} & \multicolumn{1}{c}{$\epsilon$} &  \multicolumn{1}{c}{AUC-ROC} &   \multicolumn{1}{c}{$\epsilon$}\\
\midrule
              73 &   4,381 &            0.623 &        6.117 &                  0.312 &              22.598 \\
             264 &   3,875 &            0.633 &       10.298 &                  0.472 &               0.258 \\
             420 &   3,167 &            0.580 &        1.569 &                  0.491 &               0.369 \\
             338 &   3,139 &            0.652 &        1.281 &                  0.528 &               0.376 \\
             243 &   3,026 &            0.707 &       11.217 &                  0.381 &              11.611 \\
             458 &   2,723 &            0.615 &        4.831 &                  0.289 &               4.687 \\
             167 &   2,680 &            0.727 &        4.644 &                  0.460 &               4.644 \\
             300 &   2,678 &            0.592 &       12.352 &                  0.723 &              11.121 \\
             443 &   2,666 &            0.587 &       21.097 &                  0.466 &               0.497 \\
             188 &   2,591 &            0.676 &        8.306 &                  0.501 &               7.701 \\
             208 &   2,484 &            0.542 &       17.209 &                  0.420 &              11.636 \\
             252 &   2,449 &            0.550 &        1.010 &                  0.362 &               2.274 \\
             199 &   2,215 &            0.549 &       13.064 &                  0.393 &               0.482 \\
             122 &   2,103 &            0.575 &        8.744 &                  0.458 &               1.999 \\
             176 &   1,942 &            0.591 &       15.196 &                  0.453 &               0.256 \\
             281 &   1,783 &            0.640 &        1.780 &                  0.330 &               1.321 \\
             411 &   1,747 &            0.455 &        4.298 &                  0.523 &               0.385 \\
             413 &   1,730 &            0.606 &        4.902 &                  0.592 &               2.170 \\
             449 &   1,613 &            0.440 &        0.468 &                  0.521 &               2.010 \\
             394 &   1,509 &            0.679 &        2.880 &                  0.473 &               0.700 \\
             283 &   1,478 &            0.539 &        4.307 &                  0.533 &               0.300 \\
             307 &   1,433 &            0.563 &        2.788 &                  0.344 &              17.944 \\
             331 &   1,397 &            0.483 &       17.387 &                  0.710 &               2.988 \\
             148 &   1,386 &            0.557 &        4.481 &                  0.255 &               1.633 \\
             345 &   1,372 &            0.597 &       12.883 &                  0.607 &               0.214 \\
             417 &   1,369 &            0.620 &        3.545 &                  0.606 &               4.578 \\
             165 &   1,336 &            0.667 &        4.307 &                  0.439 &               4.094 \\
             248 &   1,334 &            0.533 &        0.639 &                  0.579 &               2.596 \\
             416 &   1,330 &            0.642 &       27.024 &                  0.444 &              19.941 \\
             110 &   1,305 &            0.647 &        4.636 &                  0.337 &              11.508 \\
             183 &   1,268 &            0.554 &        0.307 &                  0.405 &               0.396 \\
\bottomrule
\end{tabular}
\end{table}
\end{document}